\documentclass[11pt]{article}

\usepackage[T1]{fontenc}
\usepackage{lmodern}
\usepackage[british]{babel}
\usepackage[margin=3cm]{geometry}
\usepackage{amsmath,amssymb,amsthm}
\usepackage[round]{natbib}
\usepackage{xcolor}
\usepackage{tikz}
\usetikzlibrary{arrows}
\usetikzlibrary{arrows.meta}
\usepackage{caption}
\usepackage{subcaption}
\usepackage[hidelinks]{hyperref}

\newtheorem{theorem}{Theorem}
\theoremstyle{definition}
\newtheorem{definition}[theorem]{Definition}

\newcommand{\sat}{\models}
\renewcommand{\phi}{\varphi}
\newcommand{\doi}[1]{\href{https://doi.org/#1}{doi:#1}}

\title{Difference-Making without Making a Difference\thanks{Accepted for publication in the \emph{Journal of Mathematical Philosophy}.}}
\author{Sander Beckers\\
University College London\\
\texttt{srekcebrednas@gmail.com}}
\date{}

\begin{document}
\maketitle

\begin{abstract} Over a series of seven papers, Andreas \& G\"unther have introduced seven definitions of actual causation and have classified them as belonging to three different, competing, types of accounts: {\em factual difference-making}, {\em counterfactual difference-making}, and {\em regularity-based}. I show that their most recent---factual difference-making---definition instantiates all three types, thereby proving that these are distinctions without a difference. I further compare their novel account to the other six accounts on several crucial examples, revealing that this undermines all seven of their accounts.  
\end{abstract}

\section{Introduction}

\citet{ag25} have recently proposed an account of actual causation, aka token causation, and compare it favourably to the accounts of several other authors by making two claims: first, it offers better verdicts on examples, and second, it replaces a muddled appeal to {\em counterfactual difference-making} with a consistent and simple appeal to {\em factual difference-making}. Yet Andreas \& G\"unther's (AG from now on) comparison to the literature contains an important oversight: AG fail to compare their novel account to the other six accounts of causation that they have themselves previously proposed ({\bf AG1--7} from now on, where {\bf AG7} is the current account). I show that comparing the current account to these earlier accounts results in falsifying both of AG’s claims, and that it undermines all seven of their accounts. 

AG partition their seven accounts into three different approaches and contrast these to each other. Informally, a {\em counterfactual} difference-making account answers the question ``did $C$ cause $E$?'' by supposing that $C$ had {\em not} occurred, and asking whether $E$ would then have failed to occur as well. A {\em factual} difference-making account instead starts from a state in which it is left open whether $C$ and $E$ occur, then supposes that $C$ {\em did} occur, and asks whether $E$ then occurs too. A {\em regularity} account, lastly, asks whether $C$ belongs to a set of actual conditions that is minimally sufficient for $E$ given the laws. Prima facie these are three different ideas, and AG take the differences to be of great philosophical importance: they argue that factual and regularity accounts are more plausible than counterfactual ones. I will argue that, at least as formalized by AG, these are distinctions without a difference. I do so by proving two results. First, the current factual difference-making account ({\bf AG7}) is equivalent to one of their most recent regularity accounts ({\bf AG5}) (which they acknowledge is itself almost identical to another recent account of theirs ({\bf AG6})). Second, I prove that there exists an equivalent formulation of their current account as a counterfactual difference-making account.

The second part of the paper concerns the examples that AG use to argue for their accounts. The upshot can be stated simply: across their papers, AG have changed their minds about what distinguishes the paradigm examples {\bf Simple Switch} and {\bf Early Preemption}, and hence about the correct verdicts on several of their variants. What makes this especially problematic for AG is that they frequently appeal to the intuitive strength of the verdicts reached on precisely these examples as evidence for the correctness of their accounts. Concretely, they are faced with an unappealing dilemma:
either they stick with their earlier view ({\bf AG1}, {\bf AG2}, {\bf AG4}) that the examples {\bf Simple Switch} and {\bf Early Preemption} are to be distinguished solely on the basis of their structural properties, in which case 
all of their later accounts offer the wrong verdict---by AG’s own lights---on several variants of the switch example that they discuss, or they stick with their more recent view ({\bf AG5}, {\bf AG6}, {\bf AG7}) 
that the distinction between these examples is due to the (non-structural) deviancy condition, in which case all of their later accounts offer the wrong verdict on cases of {\bf Early Preemption} for which the backup-process is not deviant. Additionally, choosing the second horn undermines one of the most frequent and important criticisms that AG offer of other accounts, namely that those accounts fail to structurally distinguish between {\bf Simple Switch} and {\bf Early Preemption}.

Throughout I will use the following labels to refer to the seven definitions of actual causation proposed so far by AG:
\begin{itemize}
\item {\bf AG1}:  \citet{ag20} 
\item {\bf AG2}:  \citet{ag21b} 
\item {\bf AG3}:  \citet{ag21a} 
\item {\bf AG4}: \citet{ag22}
\item {\bf AG5}: \citet{ag24a}
\item {\bf AG6}: \citet{ag24b}
\item {\bf AG7}: \citet{ag25}
\end{itemize}


\section{\texorpdfstring{Equivalence of {\bf AG7} and {\bf AG5}}{Equivalence of AG7 and AG5}}


In \citep{ag25} AG define a {\em causal model} $\langle M,V\rangle $ as consisting of a set $M$ of structural equations of the form $X= \phi$ for some binary variable $X$ and propositional formula $\phi$, and a consistent set of literals $V$. (A {\em literal} is either a variable ($X$)---representing truth---or the negation of a variable ($\lnot X$)---representing falsehood.) They also implicitly assume there to be a fixed set of variables $P$ out of which the literals in $V$ and the formulas $\phi$ can be constructed. If $V$ does not contain a literal involving $X$, then $X$ is said to be {\em unsettled}. For each variable in $P$ there is at most one equation in $M$. We say that $X$ is a {\em child} of $C$ if $C$ occurs in the equation for $X$. The {\em descendant} relation is the transitive closure of the child relation, and the {\em ancestor} relation is the reversal of the descendant relation.

The semantics of a causal model are given by combining $M$ with $V$ in the following manner. $V \sat (X = \phi)$ iff $X \in V$ and $V$ classically implies $\phi$ or $\lnot X \in V$ and $V$ classically implies $\lnot \phi$. $V \sat M$ iff $V$ satisfies each equation in $M$. $\langle M,V\rangle  \sat \phi$ for some formula $\phi$ iff $\phi$ is classically implied by $V$ and $M$.  $\langle M,V\rangle $ is {\em unsettled} about a formula $\phi$ iff $\langle M,V\rangle $ satisfies none of $\phi$ and $\lnot \phi$.\footnote{In \citep{ag21b} and \citep{ag22} they offer the same definitions but using the terminology of {\em uninformativeness} instead of {\em unsettled}.}
 
An intervention $I$ is a consistent set of literals that functions as an operator on causal models, turning $\langle M,V\rangle $ into $\langle M, V \rangle [I] := \langle M_I, V \cup I \rangle $. Here $M_I$ is defined as $M_I := \{ (X = \phi) \in M | X \not \in I \text{ and } \lnot X \not \in I\}$. Lastly, for each $X \in P$ we assume that either $X$ or $\lnot X$ is labelled as the {\em default} value, so that the opposite value can be labelled as {\em deviant}. (For now we set aside what determines this labelling.)

With these elements in place, we can present {\bf AG7}. 

\begin{definition}[{\bf Actual Causation (AG7)}]\label{def:ac} Let $\langle M,V\rangle $ be a causal model such that $V \sat M$ and $\langle M,V\rangle  \sat C \land E$. Then $C$ is a cause of $E$ relative to $\langle M,V\rangle $ iff
 there is $V' \subseteq V$ and $M' \subseteq M$ such that 
 \begin{itemize}
\item (C1) $\langle M',V'\rangle $ is unsettled on $C$ and $E$,
\item (C2) there is no $V'' \subseteq V$ such that $V' \subset V''$ and $\langle M',V''\rangle $ is unsettled on $C$ and $E$, and
\item (C3) $\langle M', \emptyset\rangle  [V'][C] \sat E$.
\item (C4) the structural equation of each descendant of $C$ is in $M'$.\footnote{Note that AG have appealed to a syntactic criterion for determining the graph structure. As a result, adding the semantically vacuous subformula 
$C \lor \lnot C$ to all equations allows one to make every variable a descendant of $C$, so that (C4) forces $M'$ to contain every equation.
This counterintuitive behaviour can be avoided by using Halpern's standard semantical criterion for the child-parent relation \citep{halpernbook}.} 
\item (C5) for any literal $C' \in V \setminus V'$ whose variable is neither a descendant nor an ancestor of $C$, $C'$ is more deviant than $\lnot C'$. 
\end{itemize}
\end{definition}

Let us walk through a simple example to illustrate {\bf AG7}. In cases of symmetric overdetermination, two independent events are both directly sufficient for producing an outcome. (Think of two soldiers shooting a victim at the same time, or less morbid, two children throwing rocks at a bottle that shatter it at the same moment.) This can be captured by the single equation causal model $E=C \lor D$, and set of literals $V=\{C,D,E\}$. The intuitive verdict here is that each of the two overdetermining events by itself is a cause of the outcome. {\bf AG7} delivers this verdict for $C$ as follows. 

Take $V'=\emptyset$ and $M'=M$. Since $M$ combined with $V'$ leaves open the truth value of both $C$ and $E$, it satisfies (C1). Furthermore, all of the sets qualifying for (C2) either contain $C$ or $E$ and are thus settled on at least one of them, or they contain $D$, which combined with $M$ settles $E$. Also, $C$ combined with $M$ implies $E$, and thus (C3) is satisfied. (C4) is trivially satisfied by taking $M'=M$. (C5) is satisfied because as a general rule, AG take truth---which usually represents an event---to be more deviant than falsehood---which usually represents an omission---in the absence of any overriding information. 

Aside from using different notation and terminology, AG define causal models and their semantics identically in \citep{ag24a} as they do in \citep{ag25}. The subsequent definition of actual causation, {\bf AG5}, differs from the current proposal only in that conditions (C1) and (C2) are replaced with the following conditions:

\begin{itemize}
\item (D1) $\langle M',V'\rangle  \not \sat E$
\item (D2) there is no $V'' \subseteq V$ such that $V' \subset V''$ and $\langle M',V''\rangle  \not \sat E$.
\end{itemize}

Yet despite this close resemblance, AG do not specify the precise relation between their novel proposal and their previous one, other than mentioning in a footnote that some of the conditions in {\bf AG7} are ``similar'' to those in {\bf AG5} \citep[132n14]{ag25}. As the following result shows, that is quite the understatement, for the two definitions are equivalent. (\citet[170]{weslake26} also remarks that both definitions consist of the ``identical formal theory,'' but without offering proof.)

\begin{theorem}\label{thm:equiv}
{\bf AG7} iff {\bf AG5}.
\end{theorem}

\begin{proof}
Given the above, it suffices to prove that in the presence of the other conditions, (C1) and (C2) are equivalent to (D1) and (D2). We refer to the preamble of the definition as (C0). Fix $V' \subseteq V$ and $M' \subseteq M$ such that (C0), (C3), (C4), and (C5) hold. We start with two observations.

First, fix some $V'' \subseteq V$ and $M'' \subseteq M$. We show that in the present setting, $\langle M'',V''\rangle $ is unsettled on $C$ and $E$  iff   $\langle M'',V''\rangle \not \sat C$ and $\langle M'',V''\rangle \not \sat E$.

Per definition, $\langle M'',V''\rangle $ is unsettled on $C$ and $E$  iff   $\langle M'',V''\rangle \not \sat C$, $\langle M'',V''\rangle \not \sat \lnot C$, $\langle M'',V''\rangle \not \sat E$, and $\langle M'',V''\rangle \not \sat \lnot E$. By the monotonicity of classical entailment, $V \sat M$ together with either $\langle M'',V''\rangle   \sat \lnot E$ or  $\langle M'',V''\rangle  \sat \lnot C$ implies that $\langle M,V\rangle  \not \sat C \land E$. (This uses the fact that $M \cup V$ is consistent, which follows from $V \sat M$: $V$ classically implies every equation in $M$, so $M \cup V$ is classically equivalent to $V$, which is a consistent set of literals.) Therefore (C0) implies that $\langle M'',V''\rangle  \not \sat \lnot E$ and $\langle M'',V''\rangle  \not \sat \lnot C$. From this the result follows.

Second, per definition, $\langle M', \emptyset\rangle  [V'][C] \sat E$ iff $\langle M'_{V' \cup \{C\}}, V' \cup \{C\}\rangle  \sat E$. (C3) and the monotonicity of classical entailment therefore imply $\langle  M',V' \cup \{C\}\rangle  \sat E$. 

We now prove the equivalence of (D1) and (C1), where the implication from right to left follows per definition. Assume (D1), that is, $\langle M',V'\rangle  \not \sat E$. Combined with $\langle  M',V' \cup \{C\}\rangle  \sat E$, we get that $\langle M',V'\rangle  \not \sat C$. Combined with our first observation, it follows that $\langle M',V'\rangle $ is unsettled on $C$ and $E$.  (Take note that one consequence of this is that without loss of generality, $V'$ may be assumed not to contain $C$, for otherwise obviously $V'$ is settled on $C$.)

It remains to prove the equivalence of (D2) and (C2), which follows from proving that for all $V'' \subseteq V$ such that $V' \subset V''$, $\langle M',V''\rangle  \not \sat E$ iff $\langle M',V''\rangle $ is unsettled on $C$ and $E$. 

Fix some $V'' \subseteq V$ such that $V' \subset V''$. We need to show that $\langle M',V''\rangle  \not \sat E$ iff $\langle M',V''\rangle $ is unsettled on $C$ and $E$. By the first observation, this reduces to showing that $\langle M',V''\rangle  \not \sat E$ iff $\langle M',V''\rangle \not \sat C$ and $\langle M',V''\rangle \not \sat E$. Therefore it suffices to show that if  $\langle M',V''\rangle  \not \sat E$ then $\langle M',V''\rangle \not \sat C$. Equivalently, it suffices to show that if $\langle M',V''\rangle  \sat C$ then $\langle M',V''\rangle  \sat E$.

Assume $\langle M',V''\rangle  \sat C$. By our second observation, we know that $\langle  M',V' \cup \{C\}\rangle  \sat E$, which implies---again by monotonicity---that $\langle  M',V'' \cup \{C\}\rangle  \sat E$. Combining these two, we get that $\langle M',V''\rangle  \sat E$, as required.
\end{proof}

Furthermore, definition {\bf AG5} is itself almost equivalent to {\bf AG6}: as \citet{ag24a} point out, the two definitions differ {\em only} regarding the precise formulation of the deviancy condition (C5). AG present (C5) as an improvement that is able to handle some counterexamples to their previous formulation, so that {\bf AG5} is strictly to be preferred over {\bf AG6}. As a result, we have that three of AG's seven definitions are essentially reduced to one. 


\section{\texorpdfstring{Differences in difference-making}{Differences in difference-making}}

AG present {\bf AG7} as a {\em factual difference-making} account of causation, whereas they present {\bf AG5} and {\bf AG6} as {\em regularity-based} accounts of causation. They remain silent on the relation between both types of accounts. As we have just shown the equivalence of {\bf AG7} and {\bf AG5}, it follows that there is in fact no difference between them. AG contrast both types of account with the more well-known {\em counterfactual difference-making} accounts of causation. In fact, this contrast forms a large part of AG's motivation for their novel accounts. Counterfactual difference-making accounts have taken over as the most popular approach to analysing causation over the past decades, to the detriment of regularity-based accounts. Well-known examples of the latter are those of \citet{mackie74} and \citet{wright88,wright11}, whereas well-known examples of the former are \citep{lewis73, halpernpearl05a, weslake, halpernbook, beckers18a, beckers21c} and many more. AG argue that the superiority of their two regularity-based accounts should make us reconsider this evolution. 

I now show that the distinction between these two kinds of accounts is also one without a difference, and thus the supposed rivalry between the two classes of approaches as they are characterized by AG is a chimera. (Elsewhere \citep{beckers21a} I make a similar point by showing how Wright's NESS (necessary element of a sufficient set) account of causation also includes a counterfactual difference-making condition. In fact, it is easy to show that {\bf AG7} satisfies an additional counterfactual condition to the NESS definition---called {\bf Counterfactual} in \citep{beckers21a}---placing it at a similar location on the spectrum between regularity accounts and counterfactual accounts as the Counterfactual NESS account.) 

Consider the following variation on {\bf AG5} (and thus on {\bf AG7}), that replaces (D1) and (D2) (resp. (C1) and (C2)) with counterfactual variants.

\begin{definition}[{\bf AG7*}]\label{def:ac2} Let $\langle M,V\rangle $ be a causal model such that $V \sat M$ and $\langle M,V\rangle  \sat C \land E$. Then $C$ is a cause of $E$ relative to $\langle M,V\rangle $ iff
 there is $V' \subseteq V$ and $M' \subseteq M$ such that 
 \begin{itemize}
\item (E1) $\langle M',V'\rangle [\lnot C] \not \sat E$
\item (E2) there is no $V'' \subseteq V$ such that $V' \subset V''$ and $\langle M',V''\rangle [\lnot C] \not \sat E$, and
\item (C3) $\langle M', \emptyset\rangle  [V'][C] \sat E$.
\item (C4) the structural equation of each descendant of $C$ is in $M'$.
\item (C5) for any literal $C' \in V \setminus V'$ whose variable is neither a descendant nor an ancestor of $C$, $C'$ is more deviant than $\lnot C'$. 
\end{itemize}
\end{definition}

Informally the difference between both definitions can be put as follows: instead of merely considering a minimal submodel in which the truth of $C$ and $E$ is unsettled, (E1) and (E2) require that we consider a minimal submodel in which the counterfactual supposition $\lnot C$ does not make $E$ true. That this is indeed a counterfactual difference-making account by AG's own lights can be seen in two ways. 

One characterization AG offer to classify a definition of causation as a counterfactual difference-making account is that it asks a counterfactual question regarding what would have happened, had the putative cause not obtained, which is precisely what occurs in (E1). Concretely, AG express the contrast as follows:
\begin{quote}
Our regularity theory does not rely on any condition of counterfactual dependence. It does {\em not} ask what would have happened, had the putative
cause not obtained. Thereby our theory does not rely on counterfactual dependence~\ldots Our regularity theory is not counterfactual. \citep[2169]{ag24b}
\end{quote}

AG also offer a second characterization, namely that counterfactual dependence is a sufficient condition for causation. Counterfactual dependence, for a model such that $V \sat M$, holds when both  $\langle M,V\rangle  \sat C \land E$ and $\langle M,V \setminus \{C\}\rangle [\lnot C]  \sat \lnot E$. If we ignore (C5), then it is trivial to see that {\bf AG7*} also satisfies this characterization: simply take $M'=M$ and $V'=V \setminus \{C\}$. Adding (C5) on the other hand prevents the sufficiency of counterfactual dependence. But the presence of a deviancy condition cannot be what separates counterfactual difference-making accounts from other types of accounts, for the most influential counterfactual account---that of \citet{halpernbook}---also comes with a deviancy condition that would preclude it from satisfying this characterization (as do many other such accounts \citep{hall07,hitchcock07,gallow21}). The only conclusion to draw is that the presence or absence of a deviancy condition stands orthogonal to the distinction between counterfactual, factual, and regularity accounts, and {\bf AG7*} is thus squarely a counterfactual difference-making account. 

Yet as the following result shows, {\bf AG7*} and {\bf AG7} are equivalent. Therefore AG's factual difference-making account is revealed to be itself an instance of a counterfactual difference-making account, as opposed to an alternative. 

\begin{theorem}\label{thm:equiv2}
{\bf AG7*} iff {\bf AG7}.
\end{theorem}

\begin{proof}
Given the above and Theorem~\ref{thm:equiv}, it suffices to prove that in the presence of the other conditions, (D1) and (D2) are equivalent to (E1) and (E2). We refer to the preamble of the definition as (C0). Fix $V' \subseteq V$ and $M' \subseteq M$ such that (C0), (C3), (C4), and (C5) hold. 

As proven by \citet{ackermans26}, for {\bf AG7} we can assume without loss of generality that $M'$ does not contain an equation for $C$. That the same holds for {\bf AG7*} is immediate, because (E1), (E2), and (C3) all intervene on $C$, so they never use $C$'s equation.

Fix some $V'' \subseteq V$ such that $V' \subseteq V''$. We need to show that $\langle M',V''\rangle [\lnot C] \not \sat E$ iff $\langle M',V''\rangle  \not \sat E$. Equivalently, $\langle M',V''\rangle [\lnot C]  \sat E$ iff $\langle M',V''\rangle  \sat E$.

We start with the implication from right to left. Assume that $\langle M',V''\rangle  \sat E$. Given that $M'$ does not contain an equation for the variable in $C$, the result is immediate by the monotonicity of classical entailment. 

Now the implication from  left to right. Assume $\langle M',V''\rangle  [\lnot C] \sat E$. By (C3), we have that $\langle M', \emptyset\rangle  [V'][C] \sat E$. By monotonicity, it follows that also $\langle M',V''\rangle [C] \sat E$. Since $M'$ has no equation for $C$, $\langle M',V''\rangle [C] = \langle M', V'' \cup \{C\}\rangle $ and $\langle M',V''\rangle [\lnot C] = \langle M', V'' \cup \{\lnot C\}\rangle $; so $M' \cup V'' \cup \{C\} \models E$ and $M' \cup V'' \cup \{\lnot C\} \models E$, which implies $M' \cup V'' \models E$, that is, $\langle M',V''\rangle  \sat E$.
\end{proof}

\citet{ackermans26} similarly proves the equivalence of {\bf AG7} to an alternative definition that has counterfactual elements. I prefer {\bf AG7*} to his variant because it is easier to interpret and compare with {\bf AG7}. I disagree with Ackermans, however, that {\bf AG7} sets itself apart from counterfactual accounts by invoking the idea of variables whose values are unspecified, for I presented the same idea in \citep{beckers21c} under the label of {\em strong sufficiency}, and it appeared already in \citep{weslake}. (Concretely, letting $N \subseteq P$ denote the variables whose equation is in $M'$, (C3) expresses that $V' \cup \{C\}$ is strongly sufficient for $E$ along the network $N$.)
  
One might object that extensional equivalence does not suffice for two accounts to be the same: definitions that agree on every verdict might still differ in what they take to be explanatorily basic. On this view, {\bf AG7} says that $C$ causes $E$ {\em because} a factual condition obtains, whereas {\bf AG7*} says so because a counterfactual condition obtains. (The worry has a precedent in \citet[565]{lewis73}, who shows that, under determinism and given an additional assumption of counterfactual independence, the conditions of a regularity analysis suffice for counterfactual dependence, and who uses this to explain why regularity analyses ``work as well as they do.'' Lewis does not take this to show that the two analyses coincide.) I grant the general point, but it does not rescue AG's taxonomy, for three reasons. 

First, the relation between {\bf AG7} and {\bf AG7*} is of an altogether different kind than the one Lewis identified. Lewis's observation is one-directional and conditional: regularity conditions suffice for counterfactual dependence only given further assumptions, and the two analyses still come apart elsewhere. Indeed, Lewis's very motivation for a counterfactual analysis is that regularity analyses go wrong on effects, epiphenomena, and preempted potential causes. By contrast, the equivalence of {\bf AG7} and {\bf AG7*} is unconditional. It is a theorem that holds across all causal models of AG's own framework, so there is no case at all on which the two definitions disagree. Moreover, both definitions are formulated in terms of exactly the same primitives---structural equations, sets of literals, interventions, and deviancy---and differ only in which entailment queries about submodels they formulate. 

Second, AG's case for preferring factual over counterfactual difference-making rests on verdicts. Counterfactual accounts are criticized for giving wrong verdicts on cases of redundant causation, or for having to give up the necessity and sufficiency of counterfactual dependence in order to get them right \citep[143--145]{ag25}. Theorem~\ref{thm:equiv2} shows that a counterfactual difference-making account delivers exactly AG's verdicts, so no argument of this kind can tell against counterfactual difference-making as such. 

Third, if the distinction is meant to concern explanatory priority rather than verdicts, then the burden is on AG to articulate what this explanatory difference consists in, and why it should bear on the evaluation of accounts of causation. Their work offers no such articulation beyond the verdicts on examples. By the standards that AG themselves employ, then, the distinction makes no difference.

In sum, AG's attempt to distinguish between regularity-based accounts, factual difference-making accounts, and counterfactual difference-making accounts has failed. 

\section{\texorpdfstring{To flip, or not to flip?}{To flip, or not to flip?}}

Throughout all of their work AG evaluate definitions of causation by comparing their verdicts on a small set of paradigmatic examples, under the assumption that there is a unique, correct, and intuitive verdict for each example. So failure to offer the right verdict on at least one of these examples is taken as a sufficient criterion for disqualifying a definition. This methodology is highly questionable, both because there is in fact no consensus on the verdicts for many of these examples (as AG admit in their more recent work), and more fundamentally, because one might prefer a more systematic foundation for causation than mere intuitions \citep{stonesoup,beckers21c,beckers24b}. I here offer a much more basic problem for AG's method: the judgments and arguments used for some of the examples in defending their later definitions directly conflict with those of their earlier definitions. As a consequence, we cannot simply interpret AG's various definitions as a sequence of modifications that steadily improve upon each other. Rather, the fact that AG changed their minds on what the correct analysis is for these paradigmatic examples undermines the very idea that definitions ought to be evaluated on the basis of what are supposed to be intuitive and universal verdicts about them. 

There are two examples of causal models in particular that AG make heavy use of: {\bf Early Preemption} and {\bf Simple Switch}. Here {\bf Early Preemption} and {\bf Simple Switch} are nothing but the labels that AG---following \citet{hall07}---give to the causal structures depicted in figures~\ref{fig:ep} and~\ref{fig:sw}, respectively. In most of their papers---\citep{ag20,ag21b,ag22,ag24a,ag24b}---they point out that other definitions fail to distinguish between them, whereas the allegedly correct verdicts on these examples are different, and they use this as an important argument in favour of their definitions. Concretely, the idea is that $C$ is a cause of $E$ in {\bf Early Preemption} but not in {\bf Simple Switch}. The only definition of theirs that does not distinguish between them and calls $C$ a cause in both is {\bf AG3} (an observation that to my knowledge has gone unnoticed so far). 

\begin{figure}
\centering
\begin{subfigure}{.5\textwidth}
\centering
\begin{tikzpicture}[node distance={20mm}, thick, main/.style = {draw, circle}] 
\node[fill=gray,main] (1) {$C$}; 
\node[fill=gray,main] (2) [above right of=1] {$D$};
\node[main] (3) [below right of=1] {$B$}; 
\node[fill=gray,main] (5) [left of=3] {$A$};
\node[fill=gray,main] (4) [below right of=2] {$E$}; 

\draw[->,>=stealth',auto] (1) -- (2);
\draw[->,>=stealth',auto] (5) -- (3);
\draw[->,>=stealth',auto] (2) -- (4);
\draw[->,>=stealth',auto] (3) -- (4);
\draw[-{Circle}] (1) -- (3);
\end{tikzpicture} 
\caption{{\bf Early Preemption}}
\label{fig:ep}
\end{subfigure}%
\begin{subfigure}{.5\textwidth}
\centering
\begin{tikzpicture}[node distance={20mm}, thick, main/.style = {draw, circle}] 
\node[fill=gray,main] (1) {$C$}; 
\node[fill=gray,main] (2) [above right of=1] {$D$};
\node[main] (3) [below right of=1] {$B$}; 
\node[fill=gray,main] (4) [below right of=2] {$E$}; 

\draw[->,>=stealth',auto] (1) -- (2);
\draw[->,>=stealth',auto] (2) -- (4);
\draw[->,>=stealth',auto] (3) -- (4);
\draw[-{Circle}] (1) -- (3);
\end{tikzpicture} 
\caption{{\bf Simple Switch}}
\label{fig:sw}
\end{subfigure}
\caption{Neuron diagrams of the two paradigmatic causal structures.}
\label{fig:neuron}
\end{figure}

The graphs in figures~\ref{fig:ep} and~\ref{fig:sw} are examples of so-called ``neuron diagrams,'' which are often used in the literature on causation to visually represent causal models. The nodes represent binary variables, and a node is grey iff the variable is set to true. The edges represent causal influence in the following manner: the receiving node is true whenever any of the edges with an arrowhead comes from a true node {\em and} there is no edge with a circlehead coming from a true node. For {\bf Early Preemption} this means that the values of the variables are determined by the following equations: $C=1$, $A=1$, $D=C$, $B=\lnot C \land A$, and lastly, $E=D \lor B$. For {\bf Simple Switch} the equations are: $C=1$, $D=C$, $B=\lnot C$, and $E=D \lor B$. 

Here is a classic scenario by \citeauthor{hitchcock01} that often accompanies {\bf Early Preemption}:
\begin{quote}
``Backup'': an assassin-in-training is on his first mission. Trainee is an
excellent shot: if he shoots his gun, the bullet will fell Victim. Supervisor
is also present, in case Trainee has a last minute loss of nerve (a common
affliction among student assassins) and fails to pull the trigger. If
Trainee does not shoot, Supervisor will shoot Victim herself. In fact,
Trainee performs admirably, firing his gun and killing Victim. \citep[276]{hitchcock01}
\end{quote}

The idea is that Trainee's shot ($C=1$) causes Victim's death ($E=1$), despite the existence of a backup process that would have resulted in the same result had Trainee not shot. The backup process consists of Supervisor's intention ($A=1$) to shoot Victim ($B$) in case Trainee does not. {\bf AG7} arrives at this conclusion by taking $V'=\{\lnot B\}$. Note that $A$ is the only variable that we have removed from $V$ which falls under the scope of (C5). In other examples of this sort, AG state that a literal is more deviant than its negation if it violates a moral norm, which is clearly the case here. So far so good. 

The standard scenario used by AG---and many others---to illustrate {\bf Simple Switch} is due to \citeauthor{hall00}:
\begin{quote}
``The Engineer'': an engineer is standing by a switch in the railroad
tracks. A train approaches in the distance. She flips the switch [$C$], so that
the train travels down the right-hand track [$D$], instead of the left [$\lnot B$]. Since
the tracks reconverge up ahead, the train arrives at its destination all
the same [$E$]; let us further suppose that the time and manner of its arrival
are exactly as they would have been, had she not flipped the switch [$\lnot C$]. \citep[205]{hall00}
\end{quote}

The majority view is that switches are not causes. That {\bf AG7} is able to respect this judgment is due to the absence of $A$, for this prevents the possibility of keeping $\lnot B \in V'$ and yet not settling $C$. 

Although there is more or less consensus on the causal judgments for these informal scenarios, views diverge widely over whether or not the causal models in figures~\ref{fig:ep} and~\ref{fig:sw} offer appropriate representations of them. At present the goal is not to take sides on this issue. What matters is that there exist two other causal models that have been proposed as alternative representations of the scenario that is captured by {\bf Simple Switch}, and as a result AG extend their analysis to show that their definitions offer the same verdicts on these representations as well. The first of these models, labelled {\bf Extended Switch} by AG \citep{ag22}, was presented by \citet{hitchcock09} to challenge the alleged difference between {\bf Early Preemption} and {\bf Simple Switch}. The second of these models---that I label {\bf Halpern Switch}---was presented by \citet{halpernbook} in order to show that his definition is capable of offering the right verdict on switch examples after all. Furthermore, \citet{halpernbook} argues that his definition also offers the right verdict on {\bf Simple Switch} once one invokes a {\em normality} condition, which is similar in spirit to the deviancy condition (C5).

AG discuss all three representations in their work, but in very different manners. Initially AG focus solely on the contrast between {\bf Early Preemption} and {\bf Simple Switch}, and do so relying solely on the structural features of these models \citep{ag20,ag21b}. \citet{ag22} then add a discussion of the other variants as well. There they criticize Halpern's appeal to normality considerations for handling {\bf Simple Switch}, and show that their definition ({\bf AG4}) arrives at the same verdict on all three switch representations by relying only on structural features of the models. This approach is abandoned in \citep{ag24a,ag24b}. Instead, there they show that {\bf AG5} and {\bf AG6} require invoking their non-structural deviancy condition in order for the verdicts on the two alternative switches to agree with {\bf Simple Switch}. On this latter view, therefore, cases of {\bf Early Preemption} and cases of switches {\em cannot} be distinguished purely on structural grounds. As a result, AG's defence of their recent accounts directly conflicts with the alleged clear-cut structural distinction between {\bf Early Preemption} and {\bf Simple Switch} that played such an important role in their earlier accounts.

We can make this more concrete by considering a variation of {\bf Early Preemption} in which the deviancy conditions differ. Imagine our Backup story continues as follows.

\begin{quote}``Backup 2'': We spoke too soon: Victim did not die, but their life is hanging by a thread. A doctor-in-training is on her first residency as our gunshot Victim enters the emergency room. Trainee is an
excellent surgeon: if she operates, the bullet will be successfully removed and Victim survives. Supervisor
is also present, in case Trainee has a last minute loss of nerve (a common
affliction among student doctors) and fails to operate. If
Trainee does not operate, Supervisor will operate on Victim himself. In fact,
Trainee performs admirably, removing the bullet and saving Victim's life.
\end{quote}

Both backup scenarios are structurally identical. Plausibly, so are the most intuitive verdicts for both of them: trainee causes the outcome (death or survival) in both. Yet in the latter scenario, supervisor's intention {\em is}  in accordance with a moral norm, and their failure to supervise the operation would be in violation of one. Therefore {\bf AG5}, {\bf AG6}, and {\bf AG7} force us to conclude that trainee's action did not cause the outcome in Backup 2. 

Regardless of what one makes of these opposing verdicts in isolation, the general point is simple. AG's verdicts on {\bf Simple Switch}, {\bf Early Preemption}, and their variants have not been stable: what they earlier took to be a structural difference between these examples, they now attribute to deviancy, with the result that their later accounts contradict verdicts that their earlier accounts were designed to secure. Intuitions that their own proponents have revised in this way cannot serve as fixed and widely shared data. Hence the fact that {\bf AG7} respects AG's current verdicts provides little support for it, and AG's flip-flopping makes it very hard to identify a persuasive motivation for any one of their seven accounts.

\section{Conclusion}

Over a series of seven papers, Andreas \& G\"unther have introduced seven definitions of actual causation using several variations of definitions of causal models and the tools that we equip them with. 
Throughout their work they present their definitions as belonging to competing types of accounts of causation, culminating in the contrast between (counter)factual difference-making and regularity-based accounts. I have shown that their most recent addition ({\bf AG7}) does not in fact make a difference (as it is equivalent to an earlier definition ({\bf AG5})), that the distinction between these three types of accounts also does not in fact make a difference (as {\bf AG7} can be interpreted as an instance of all three of them), and that the change in their analysis of two key examples undermines the coherence of their work as a whole.


\section*{Acknowledgements}
I thank the reviewers of this journal for their efficient and insightful reviews. This work was funded by CHAI---the EPSRC Causality in Healthcare AI Hub (grant no.\ EP/Y028856/1).


\bibliographystyle{plainnat}
\bibliography{AG_criticism_arXiv}

\begin{thebibliography}{27}
\providecommand{\natexlab}[1]{#1}
\providecommand{\url}[1]{\texttt{#1}}
\expandafter\ifx\csname urlstyle\endcsname\relax
  \providecommand{\doi}[1]{doi: #1}\else
  \providecommand{\doi}{doi: \begingroup \urlstyle{rm}\Url}\fi

\bibitem[Ackermans(2025)]{ackermans26}
Lennart~B. Ackermans.
\newblock Factual difference-making is equivalent to a counterfactual theory.
\newblock \emph{Australasian Philosophical Review}, 9\penalty0 (2):\penalty0
  205--212, 2025.
\newblock \doi{10.1080/24740500.2025.2592582}.

\bibitem[Andreas and G{\"u}nther(2020)]{ag20}
Holger Andreas and Mario G{\"u}nther.
\newblock Causation in terms of production.
\newblock \emph{Philosophical Studies}, 177\penalty0 (6):\penalty0 1565--1591,
  2020.
\newblock \doi{10.1007/s11098-019-01275-3}.

\bibitem[Andreas and G{\"u}nther(2021{\natexlab{a}})]{ag21a}
Holger Andreas and Mario G{\"u}nther.
\newblock A {Ramsey} test analysis of causation for causal models.
\newblock \emph{British Journal for the Philosophy of Science}, 72\penalty0
  (2):\penalty0 587--615, 2021{\natexlab{a}}.
\newblock \doi{10.1093/bjps/axy074}.

\bibitem[Andreas and G{\"u}nther(2021{\natexlab{b}})]{ag21b}
Holger Andreas and Mario G{\"u}nther.
\newblock Difference-making causation.
\newblock \emph{Journal of Philosophy}, 118\penalty0 (12):\penalty0 680--701,
  2021{\natexlab{b}}.
\newblock \doi{10.5840/jphil20211181243}.

\bibitem[Andreas and G{\"u}nther(2022)]{ag22}
Holger Andreas and Mario G{\"u}nther.
\newblock Actual causation.
\newblock \emph{Dialectica}, 76\penalty0 (1):\penalty0 1--44, 2022.
\newblock \doi{10.48106/dial.v76.i1.01}.

\bibitem[Andreas and G{\"u}nther(2024{\natexlab{a}})]{ag24a}
Holger Andreas and Mario G{\"u}nther.
\newblock A regularity theory of causation.
\newblock \emph{Pacific Philosophical Quarterly}, 105\penalty0 (1):\penalty0
  2--32, 2024{\natexlab{a}}.
\newblock \doi{10.1111/papq.12447}.

\bibitem[Andreas and G{\"u}nther(2024{\natexlab{b}})]{ag24b}
Holger Andreas and Mario G{\"u}nther.
\newblock A {Lewisian} regularity theory.
\newblock \emph{Philosophical Studies}, 181:\penalty0 2145--2176,
  2024{\natexlab{b}}.
\newblock \doi{10.1007/s11098-024-02149-z}.

\bibitem[Andreas and G{\"u}nther(2025)]{ag25}
Holger Andreas and Mario G{\"u}nther.
\newblock Factual difference-making.
\newblock \emph{Australasian Philosophical Review}, 9\penalty0 (2):\penalty0
  118--148, 2025.
\newblock \doi{10.1080/24740500.2025.2477315}.

\bibitem[Beckers(2021{\natexlab{a}})]{beckers21a}
Sander Beckers.
\newblock The counterfactual {NESS} definition of causation.
\newblock In \emph{Proceedings of the {AAAI} Conference on Artificial
  Intelligence}, volume~35, pages 6210--6217, 2021{\natexlab{a}}.
\newblock \doi{10.1609/aaai.v35i7.16772}.

\bibitem[Beckers(2021{\natexlab{b}})]{beckers21c}
Sander Beckers.
\newblock Causal sufficiency and actual causation.
\newblock \emph{Journal of Philosophical Logic}, 50:\penalty0 1341--1374,
  2021{\natexlab{b}}.
\newblock \doi{10.1007/s10992-021-09601-z}.

\bibitem[Beckers(2025)]{beckers24b}
Sander Beckers.
\newblock Actual causation and nondeterministic causal models.
\newblock In Biwei Huang and Mathias Drton, editors, \emph{Proceedings of the
  Fourth Conference on Causal Learning and Reasoning}, volume 275 of
  \emph{Proceedings of Machine Learning Research}, pages 514--532. PMLR, 2025.
\newblock URL \url{https://proceedings.mlr.press/v275/beckers25a.html}.

\bibitem[Beckers and Vennekens(2018)]{beckers18a}
Sander Beckers and Joost Vennekens.
\newblock A principled approach to defining actual causation.
\newblock \emph{Synthese}, 195\penalty0 (2):\penalty0 835--862, 2018.
\newblock \doi{10.1007/s11229-016-1247-1}.

\bibitem[Gallow(2021)]{gallow21}
J.~Dmitri Gallow.
\newblock A model-invariant theory of causation.
\newblock \emph{Philosophical Review}, 130\penalty0 (1):\penalty0 45--96, 2021.
\newblock \doi{10.1215/00318108-8699682}.

\bibitem[Glymour et~al.(2010)Glymour, Danks, Glymour, Eberhardt, Ramsey,
  Scheines, Spirtes, Teng, and Zhang]{stonesoup}
Clark Glymour, David Danks, Bruce Glymour, Frederick Eberhardt, Joseph Ramsey,
  Richard Scheines, Peter Spirtes, Choh~Man Teng, and Jiji Zhang.
\newblock Actual causation: A stone soup essay.
\newblock \emph{Synthese}, 175\penalty0 (2):\penalty0 169--192, 2010.
\newblock \doi{10.1007/s11229-009-9497-9}.

\bibitem[Hall(2000)]{hall00}
Ned Hall.
\newblock Causation and the price of transitivity.
\newblock \emph{Journal of Philosophy}, 97\penalty0 (4):\penalty0 198--222,
  2000.
\newblock \doi{10.2307/2678390}.

\bibitem[Hall(2007)]{hall07}
Ned Hall.
\newblock Structural equations and causation.
\newblock \emph{Philosophical Studies}, 132\penalty0 (1):\penalty0 109--136,
  2007.
\newblock \doi{10.1007/s11098-006-9057-9}.

\bibitem[Halpern(2016)]{halpernbook}
Joseph~Y. Halpern.
\newblock \emph{Actual Causality}.
\newblock MIT Press, 2016.

\bibitem[Halpern and Pearl(2005)]{halpernpearl05a}
Joseph~Y. Halpern and Judea Pearl.
\newblock Causes and explanations: A structural-model approach. {Part I}:
  Causes.
\newblock \emph{British Journal for the Philosophy of Science}, 56\penalty0
  (4):\penalty0 843--887, 2005.
\newblock \doi{10.1093/bjps/axi147}.

\bibitem[Hitchcock(2001)]{hitchcock01}
Christopher Hitchcock.
\newblock The intransitivity of causation revealed in equations and graphs.
\newblock \emph{Journal of Philosophy}, 98\penalty0 (6):\penalty0 273--299,
  2001.
\newblock \doi{10.2307/2678432}.

\bibitem[Hitchcock(2007)]{hitchcock07}
Christopher Hitchcock.
\newblock Prevention, preemption, and the principle of sufficient reason.
\newblock \emph{Philosophical Review}, 116\penalty0 (4):\penalty0 495--532,
  2007.
\newblock \doi{10.1215/00318108-2007-012}.

\bibitem[Hitchcock(2009)]{hitchcock09}
Christopher Hitchcock.
\newblock Structural equations and causation: Six counterexamples.
\newblock \emph{Philosophical Studies}, 144\penalty0 (3):\penalty0 391--401,
  2009.
\newblock \doi{10.1007/s11098-008-9216-2}.

\bibitem[Lewis(1973)]{lewis73}
David Lewis.
\newblock Causation.
\newblock \emph{Journal of Philosophy}, 70\penalty0 (17):\penalty0 556--567,
  1973.
\newblock \doi{10.2307/2025310}.

\bibitem[Mackie(1974)]{mackie74}
J.~L. Mackie.
\newblock \emph{The Cement of the Universe}.
\newblock Oxford University Press, 1974.

\bibitem[Weslake(2025)]{weslake26}
Brad Weslake.
\newblock Commentary on factual difference-making.
\newblock \emph{Australasian Philosophical Review}, 9\penalty0 (2):\penalty0
  167--176, 2025.
\newblock \doi{10.1080/24740500.2025.2592584}.

\bibitem[Weslake(forthcoming)]{weslake}
Brad Weslake.
\newblock A partial theory of actual causation.
\newblock \emph{British Journal for the Philosophy of Science}, forthcoming.

\bibitem[Wright(1988)]{wright88}
Richard~W. Wright.
\newblock Causation, responsibility, risk, probability, naked statistics, and
  proof: Pruning the bramble bush by clarifying the concepts.
\newblock \emph{Iowa Law Review}, 73:\penalty0 1001--1077, 1988.

\bibitem[Wright(2011)]{wright11}
Richard~W. Wright.
\newblock The {NESS} account of natural causation: A response to criticisms.
\newblock In Richard Goldberg, editor, \emph{Perspectives on Causation}, pages
  285--322. Hart Publishing, 2011.

\end{thebibliography}

\end{document}